\documentclass[letterpaper, 10pt, conference]{ieeeconf}

\IEEEoverridecommandlockouts   
\usepackage{newtxtext}   
\usepackage{microtype}
\usepackage{graphicx}
\usepackage{booktabs}
\usepackage{multirow}
\usepackage{amsmath,amssymb}
\usepackage{cite}   
\usepackage{hyperref}
\usepackage[table]{xcolor}
\usepackage{caption}
\usepackage{tabularx}
\usepackage{subcaption}

\title{\LARGE \bf BIND: Binding 3D Robot Actions\\to 2D Image Features}

\newif\ifanon
\anonfalse
\ifanon
\author{Anonymous Authors}
\else
\author{%
  Cameron Smith$^{1}$, Arsh Tangri$^{2}$, Vitor Guizilini$^{2}$, Yue Wang$^{1}$, Zubair Irshad$^{2}$, Sergey Zakharov$^{2}$
  \thanks{$^{1}$University of Southern California}%
  \thanks{$^{2}$Toyota Research Institute}%
}
\fi

\makeatletter
\def\bstctlcite{\@ifnextchar[{\@bstctlcite}{\@bstctlcite[@auxout]}}
\def\@bstctlcite[#1]#2{\@bsphack
  \@for\@citeb:=#2\do{%
    \edef\@citeb{\expandafter\@firstofone\@citeb}%
    \if@filesw\immediate\write\csname #1\endcsname{\string\citation{\@citeb}}\fi}%
  \@esphack}
\makeatother

\begin{document}
\bstctlcite{BSTcontrol}
\maketitle
\thispagestyle{empty}
\pagestyle{empty}

\begin{abstract}
We introduce \textbf{BIND}, a new action representation for visuomotor robot policies that binds 3D robot actions to their corresponding 2D image features, yielding strong data efficiency gains and robustness to out-of-distribution object positions and camera viewpoints.
The action heads of current robot policies are typically formulated as an MLP regression from a single global feature vector produced by a pretrained vision encoder.
This global formulation requires the policy network to discover, from demonstrations alone, the relationship between target robot actions and the image features they project onto.
The consequence is that although modern image features are semantically descriptive, spatially robust, and even multiview-consistent, the policies built on them are brittle to subtle changes in camera viewpoint and object placement---and surprisingly data-inefficient.
BIND closes this gap by supplying the action--feature relationship through camera geometry rather than learning: it discretizes a volume of candidate end effector positions, attaches each candidate to the pretrained features at its projection in each camera view, and selects actions by scoring each candidate's position and image-bound feature combination.
On a real robot, we study data efficiency and out-of-distribution robustness to unseen object positions and camera viewpoints, as well as general long-horizon task execution and dexterity.
We find BIND to be highly data-efficient and robust: it achieves near-perfect success on tasks with as few as 5 demonstrations, and degrades gracefully under steep camera-viewpoint shifts and held-out object positions where coordinate-regression baselines completely fail.
\end{abstract}

\begin{figure*}[t!]
\centering
\includegraphics[width=\textwidth]{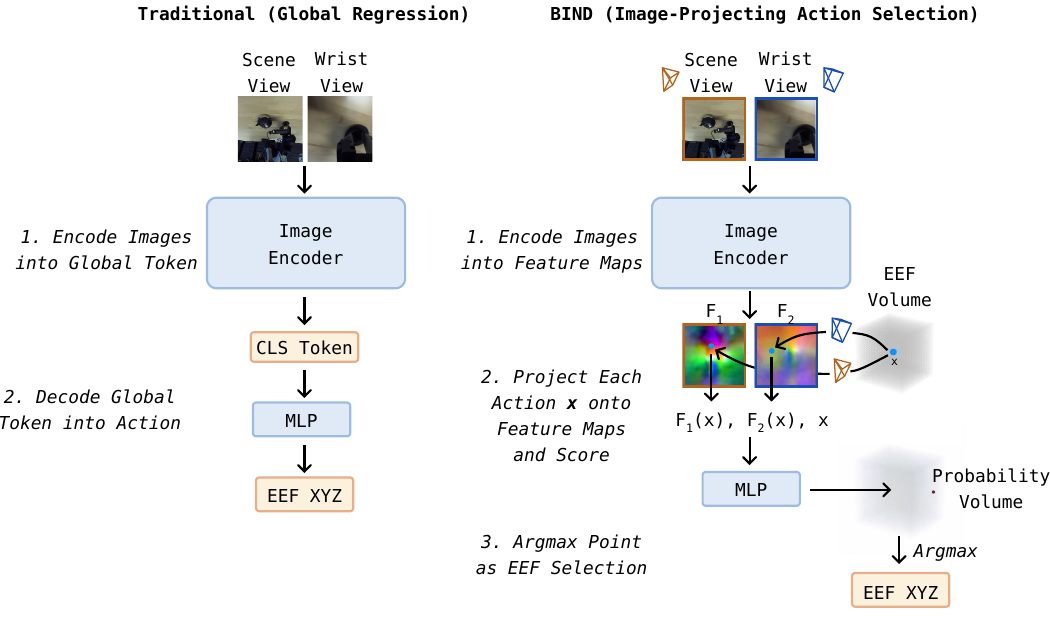}
\vspace{-0.2in}
\caption{\textbf{Traditional global regression vs.\ BIND projection-anchored selection.}
Standard visuomotor policies (left) pool image features into a global descriptor and regress robot actions---forcing the network to discover the robot-action$\leftrightarrow$pixel mapping from demonstrations.
BIND (right) instead discretizes a volume of candidate end effector positions and \emph{binds} each candidate to the image features at its projection in each camera view via known camera poses and intrinsics. 
Intuitively, this offers the policy the information of `choosing this candidate robot action would move the robot EEF to this image feature.'}
\label{fig:overview}
\end{figure*}

\section{Introduction}
\label{sec:intro}

Behavioral cloning---training a robot policy as a supervised mapping from observations to actions on demonstration data---is typically instantiated by adding an MLP head that regresses robot actions from the globally pooled feature vector of a pretrained image encoder.
The encoders themselves are strong: recent studies demonstrate that modern image features are semantically descriptive, spatially consistent, and even multiview-consistent, and that simple zero-shot probes recover geometry and correspondence from them.
The policies built on these features, however, remain brittle even on simple tasks: small changes in camera pose, lighting, or object placement relative to the training distribution frequently cause complete task failure, despite the underlying features being largely invariant to such shifts.
Robot policies, in other words, do not easily inherit the robustness of their features.
We attribute this gap in large part to the fact that the policy network must discover, from demonstrations alone, the \emph{projective} relationship between robot actions and their projecting image features: for the policy to command the robot to reach a point observed in the image, it must implicitly learn which actions would place the end effector there.
However, this relationship between the end effector and its projecting image features is uniquely and simply determined by camera geometry: for an end effector's position $\mathbf{x}$, the corresponding image feature lies exactly at the projection $\pi(\mathbf{x})$ given by the camera pose and intrinsics.
Robot policies today are typically trained or fine-tuned on demonstrations collected from a single viewpoint, over a narrow object distribution, and in modest quantity---often yielding insufficient coverage to disentangle the projective relationship from the joint distribution over camera poses, object positions, scenes, and actions.
The mapping from a 3D robot action to its corresponding image features is a single, closed-form projection equation determined by the camera pose and intrinsics, and need not be learned at all; our work instead supplies the policy with this projective relationship directly.

BIND discretizes a volume of candidate end effector positions: each 3D action candidate is attached with its projecting image features in every camera view to form an action--feature pair.
Each action--feature pair is then simply scored in typical classification-supervision style (Fig.~\ref{fig:overview}). We find that by making this relationship between robot actions and corresponding image features explicit, policies benefit from strong data-efficiency gains and robustness to out-of-distribution object positions and camera viewpoints. 

\begin{figure*}[t!]
\centering
\includegraphics[width=\textwidth]{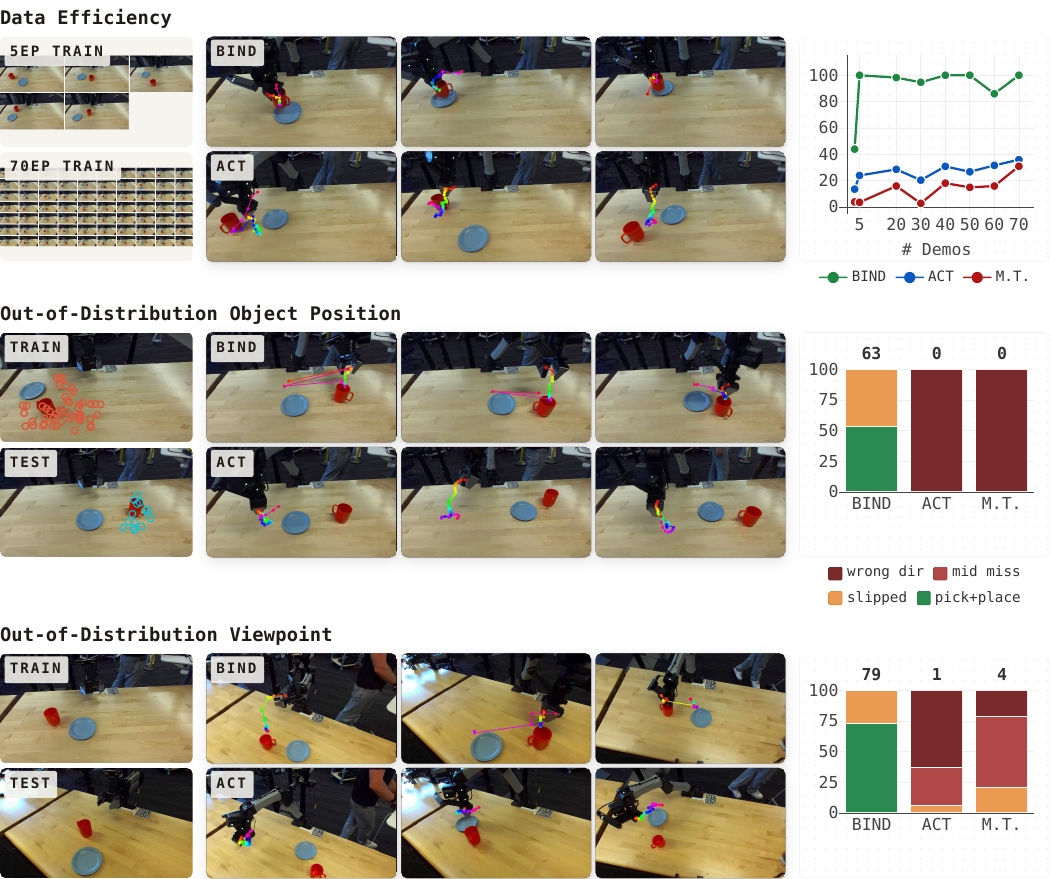}
\vspace{-0.2in}
\caption{\textbf{Data Efficiency and Out-of-Distribution Experiments: BIND vs.\ global-regression (ACT) and Motion-Tracks baselines.}
Across data efficiency (top), OOD object positions (middle), and OOD camera viewpoints (bottom), BIND's projection-anchored robot action selection outperforms the global-regression ACT and Motion Tracks baselines.
All heads share the same DINOv3 backbone and training data---the only difference is the action head.}
\label{fig:real_results}
\end{figure*}

\begin{figure*}[t!]
\centering
\includegraphics[width=\textwidth]{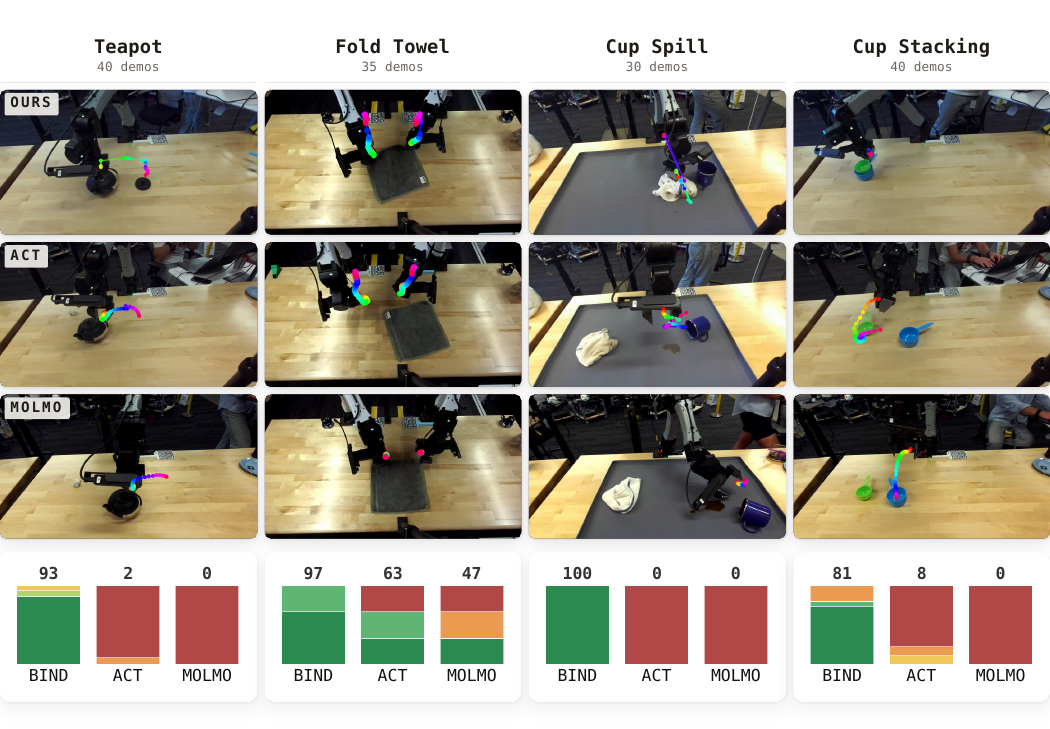}
\vspace{-0.2in}
\caption{\textbf{Long-horizon and dexterous real-robot tasks.}
Rollouts across four tasks (left to right): teapot preparation, towel folding, spill cleanup, and cup stacking.
Rows compare BIND against the ACT global-regression baseline and a fully-finetuned Molmo VLA; per-task bars report success rates for each method.}
\label{fig:other_tasks}
\end{figure*}

\section{Related Work}
\label{sec:related}

\paragraph{Imitation learning}
Imitation learning aims to train robot policies from demonstration data~\cite{schaal1996lfd,atkeson1997lfd,levine2016visuomotor,torabi2018bco,duan2017oneshot}: a human typically demonstrates behavior (often via teleoperation or hand-held data collection devices~\cite{wu2023gello,chi2024umi}), and a visuomotor policy is supervised to produce the same actions given input observations.
In its modern form, demonstrations directly serve as supervised data to learn a mapping from image observations to robot actions, without hand-designed state estimation or motion planning in between.

\paragraph{Global-regression policies}
The dominant instantiation of this mapping is what we term \emph{global regression}: image features are pooled into a single global vector, and an action head MLP decodes robot actions from it.
ACT~\cite{zhao2023aloha} regresses action chunks from a vision transformer's CLS token with an MSE objective; Diffusion Policy~\cite{chi2023diffusionpolicy} replaces direct coordinate regression with a denoising head; $\pi_0$~\cite{black2024pi0} and its successors decode actions with flow matching.
Recent video-generation-based policy architectures likewise attach action heads that decode from a global representation alongside the predicted video~\cite{du2023unipi,li2025uva,kim2026cosmospolicy,ye2026dreamzero}.
Architecturally, earlier models instantiated the global pooling as the final spatial pooling layer of a CNN feature map, while modern methods use vision transformers' dedicated CLS token or append a dedicated `action token'.
Although these heads differ in their supervising objective---MSE, diffusion, or flow matching---they share the interface of globally pooling image features into a global token: the association between actions and the image features that ground them must still be discovered from demonstration data.

\paragraph{Voxel-categorical translation heads}
PerAct~\cite{shridhar2022peract}, C2F-ARM~\cite{james2022c2farm}, and GNFactor~\cite{ze2023gnfactor} predict end effector translation as a categorical distribution over a 3D voxel grid.
This action head is similar in spirit to ours in binding robot actions to the 3D points they occupy, some of which intersect observed scene surfaces.
The binding, however, is mediated by a heavyweight perception stack: RGB-D observations are voxelized into a fused 3D representation---only voxels at the surface offer image-lifted features (or often even just RGB), while the free-space voxels that most of a trajectory passes through offer none.
Partly as a consequence, these methods typically predict sparse keyframe or waypoint poses rather than temporally dense actions, since the free-space trajectories offer little pretrained learning signal.
BIND retains the discretize-and-classify formulation but binds \emph{every} candidate to image features by projection (a free-space candidate still receives the features of the scene content it projects onto), and predicts dense-in-time action chunks.
Our method also importantly does not require depth sensors and instead operates on posed-RGB images.
These pipelines also often retrain their entire perception stack in the voxelized 3D space, forgoing pretrained 2D encoders and the internet-scale knowledge they embed; BIND instead fully leverages the image backbones themselves and also fully fine-tunes them jointly.

\paragraph{Rendered-view heatmap methods}
RVT~\cite{goyal2023rvt} and RVT-2~\cite{goyal2024rvt2} also start from 3D point clouds but, rather than voxelizing them, re-render the cloud into virtual views and apply a pretrained vision encoder to the renders.
The motivation behind these methods is to re-render the scene content from a fixed viewpoint to approximate being a camera-viewpoint-invariant policy.
They similarly bind image features to robot actions but require a cumbersome multi-view triangulation step: an action head predicts a 2D heatmap in each view, and the 3D action is then recovered by triangulating the heatmaps on the multi-view renders.
Also like the voxel-categorical methods above, these models typically output sparse keyframe waypoints.
The perception stack also remains heavyweight and clunky, requiring multi-view depth sensors and point-cloud re-rendering.
BIND reverses the direction of the binding: rather than predicting in 2D and lifting to 3D, it enumerates all candidate actions in 3D and binds each one \emph{down} to its 2D image features---avoiding triangulation, re-rendering, and depth sensing.

\paragraph{Affordance and dense pick-and-place models}
Transporter Networks~\cite{zeng2020transporter} and CLIPort~\cite{shridhar2021cliport} produce dense pick-and-place heatmaps and similarly bind robot actions to image features but are limited to top-down planar tasks.
Affordance models~\cite{tang2025uad,nasiriany2024rtaffordance,kuang2024ram} predict a single static interaction heatmap per image, often used as implicit guidance for downstream global-regression action models.
BIND predicts full multi-step 6-DoF trajectories as feature selection over a candidate volume for each timestep, rather than a single grasp point.

\section{Method}
\label{sec:method}

\begin{figure*}[t!]
\centering
\includegraphics[width=\textwidth]{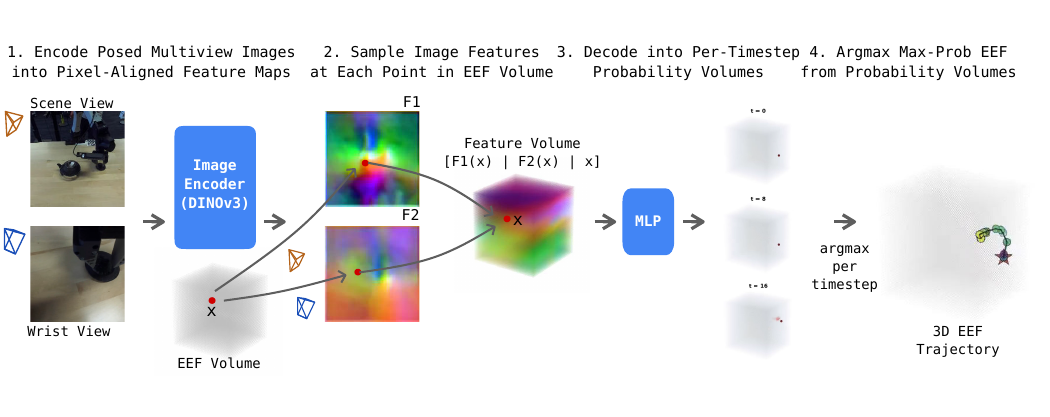}
\vspace{-0.2in}
\caption{\textbf{BIND forward pass.}
(1)~An image encoder (DINOv3) encodes the posed scene and wrist views into pixel-aligned feature maps $F_1, F_2$.
(2)~Each candidate robot action EEF point $\mathbf{x}$ in the discretized EEF volume is projected to its image features from both views, forming the action-feature pair $[F_1(\mathbf{x}) \,|\, F_2(\mathbf{x}) \,|\, \mathbf{x}]$.
(3)~An MLP decodes each action-position combination into per-timestep probability volumes.
(4)~The per-timestep argmax over each probability volume yields the maximum-probability 3D EEF trajectory.}
\label{fig:method_pipeline}
\end{figure*}

BIND transforms the standard global-regression action head into an image-feature-aware action head in two conceptual steps. (Fig.~\ref{fig:method_pipeline} gives a step-by-step walkthrough of the full pipeline.
First, we \emph{discretize} a space of robot actions: we discretize a world-space volume of candidate end effector positions and cast action prediction as classification over these candidates.
Second, we \emph{bind}: before scoring each candidate robot action, we project each candidate 3D action onto every camera view to sample its corresponding image features. We concatenate the candidate position and image feature and a small classifier network scores action--feature pairs.

\subsection{Problem Setup}
We consider behavioral cloning from demonstrations $\mathcal{D} = \{(I^{(1)}_t, \ldots, I^{(K)}_t, a_t, \{K_k\}, \{T^{\text{r}\to\text{cam}}_k\})\}$, where each $I^{(k)}_t$ is an RGB view, $a_t = (p_t \in \mathbb{R}^3, R_t \in \mathrm{SO}(3), g_t \in \mathbb{R})$ is the end effector action, and $K_k$, $T^{\text{r}\to\text{cam}}_k$ are the per-view intrinsic and robot-to-camera extrinsic matrices.
The policy predicts the next $T$ dense-in-time end effector actions from the current observation.

\subsection{Step 1: Discretizing the Action Space}
\label{sec:discretize}

Discretizing the entire robot action space (e.g. joint positions) is computationally intractable. Instead, we choose to discretize a volume of candidate end effector positions, and predict rotation and gripper separately (Section~\ref{sec:rotgrip}.
While many ways exist to construct a volume of candidate locations, in practice, we construct the volume by unprojecting each pixel from each view on a lower-resolution image grid with a number of discretized height positions along each pixel's ray.

End effector classification is predicted and supervised in a standard token-classification setup:
a network predicts a $\ell_{t,v}$ for each future timestep $t$ (Sections~\ref{sec:binding} and~\ref{sec:scoring}); training minimizes cross-entropy against the candidate nearest the ground-truth position,

\begin{equation}
\mathcal{L}_{\text{trans}} = -\frac{1}{T}\sum_{t=1}^{T} \log \frac{\exp \ell_{t,v^*_t}}{\sum_v \exp \ell_{t,v}},
\quad
v^*_t = \arg\min_v\, \|\mathbf{x}_v - p_t\|,
\end{equation}
and inference takes the argmax, $\hat{v}_t = \arg\max_v\, \ell_{t,v}$, with $\hat{p}_t = \mathbf{x}_{\hat{v}_t}$.

\subsection{Step 2: Binding Actions to Image Features by Projection}
\label{sec:binding}

For each candidate action at position $\mathbf{x} \in \mathbb{R}^3$ and each view $k$, we know the corresponding pixel locations via calibrated robot-camera pose and intrinsics.
\begin{equation}
\mathbf{u}^{(k)}(\mathbf{x}) \;=\; \pi\!\left( K_k \, T^{\text{r}\to\text{cam}}_k \, \mathbf{x} \right).
\end{equation}
To obtain the corresponding image feature, we bilinearly sample the view's spatial feature map $F_k$ at $\mathbf{u}^{(k)}(\mathbf{x})$.

\begin{table}[t!]
\centering
\setlength{\tabcolsep}{2pt}
\caption{RoboTwin success rate (\%) across five tasks.
$^{\ast}$ManiFlow rows are quoted from their paper~\cite{maniflow2025}; our evaluation modifies the simulator viewpoint, so those rows are not strictly matched to ours.
ManiFlow-3D consumes point clouds; ManiFlow-2D, ACT, and Diffusion Policy use RGB and proprioception.}
\label{tab:robotwin}
\begin{tabular*}{\linewidth}{@{\extracolsep{\fill}}lcccccc}
\toprule
Method & Dual B. & Div. B. & Shoe & Cup & Apple & Mean \\
\midrule
ManiFlow-3D$^{\ast}$ & 54.0 & 72.3 & 68.3 & 72.7 & 42.0 & 61.9 \\
ManiFlow-2D$^{\ast}$ & 47.3 & 37.0 & 45.3 & 63.7 & 37.3 & 46.1 \\
ACT                  & 32.0 & 5.0  & 29.0 & 63.0 & 42.0 & 34.2 \\
Diff.\ Policy        & 30.0 & 11.0 & 18.0 & 36.0 & 40.0 & 27.0 \\
\rowcolor{gray!15}
\textbf{BIND (ours)} & \textbf{97.0} & 66.2 & 63.0 & \textbf{77.3} & \textbf{85.4} & \textbf{77.8} \\
\bottomrule
\end{tabular*}
\end{table}

\begin{table}[t!]
\centering
\setlength{\tabcolsep}{4pt}
\caption{\textbf{Real-robot cup pick-and-place.}}
\label{tab:results}
\label{tab:data_efficiency}
\label{tab:real_ood}
\label{tab:real_oodview}
\begin{tabular}{lccc}
\toprule
 & BIND (ours) & ACT & Motion Tracks \\
\midrule
\multicolumn{4}{l}{\emph{a) Data efficiency: score vs.\ training-set size}} \\
\quad 3 demos  & \textbf{44}   & 14  & 4 \\
\quad 5  & \textbf{100} & 24  & 4 \\
\quad 20 & \textbf{98}  & 29  & 16 \\
\quad 30 & \textbf{95}  & 21  & 3  \\
\quad 40 & \textbf{100} & 31 & 18 \\
\quad 50 & \textbf{100} & 27  & 15 \\
\quad 60 & \textbf{86}   & 32  & 16 \\
\quad 70 & \textbf{100} & 36 & 31 \\
\midrule
\multicolumn{4}{l}{\emph{b) Zero-shot OOD object position: mean progress}} \\
\quad  & \textbf{63} & 0 & 0 \\
\midrule
\multicolumn{4}{l}{\emph{c) Zero-shot OOD viewpoint (3 views pooled): mean progress}} \\
\quad  & \textbf{79} & 1 & 4 \\
\bottomrule
\end{tabular}
\end{table}

\begin{table}[t!]
\centering
\setlength{\tabcolsep}{4pt}
\caption{LIBERO success rate (\%) on the Spatial and Object suites. Rows sorted by the mean of the two suites.}
\label{tab:libero}
\begin{tabular}{lccc}
\toprule
Model & Spatial & Object & Avg \\
\midrule
TraceVLA          & 84.6 & 85.2 & 84.9 \\
OpenVLA           & 84.7 & 88.4 & 86.6 \\
SpatialVLA        & 88.2 & 89.9 & 89.1 \\
CoT-VLA           & 87.5 & 91.6 & 89.6 \\
ThinkAct          & 88.3 & 91.4 & 89.9 \\
MolmoAct-7B-D     & 87.0 & 95.4 & 91.2 \\
\rowcolor{gray!15}
\textbf{BIND (ours)} & 92.0 & 94.0 & 93.0 \\
NORA-1.5          & 97.3 & 96.4 & 96.9 \\
GR00T~N1.7        & 97.7 & 97.5 & 97.6 \\
$\pi_0$           & 96.8 & 98.8 & 97.8 \\
$\pi_{0.5}$       & 98.8 & 98.2 & 98.5 \\
MolmoAct2         & 97.8 & 100.0 & 98.9 \\
MolmoAct2-Think   & 98.8 & 99.8 & 99.3 \\
\bottomrule
\end{tabular}
\end{table}

\begin{table}[t!]
\centering
\setlength{\tabcolsep}{4pt}
\caption{\textbf{Controlled simulation (MuJoCo cube pick-and-place).}
Cells report mean progress (0--100, partial credit for touching, picking, placing, and centering the cube) / binary success~\%; $n{=}50$ episodes for in-distribution and OOD object position, $n{=}196$ held-out cameras for OOD viewpoint (single training camera).}
\label{tab:sim}
\begin{tabular}{lccc}
\toprule
 & In-dist. & OOD obj. & OOD view \\
\midrule
\rowcolor{gray!15}
\textbf{BIND (ours)} & \textbf{94\,/\,94} & \textbf{96\,/\,96} & \textbf{49\,/\,44} \\
Diffusion-x0  & 82\,/\,80 & 0\,/\,0 & 12\,/\,10 \\
MolmoAct2     & 81\,/\,80 & 0\,/\,0 & 12\,/\,11 \\
ACT           & 45\,/\,46 & 0\,/\,0 & 12\,/\,7 \\
DP3           & 23\,/\,6  & 0\,/\,0 & 8\,/\,3 \\
Motion Tracks & 12\,/\,2  & 0\,/\,0 & 4\,/\,2 \\
\bottomrule
\end{tabular}
\end{table}

\begin{table}[t!]
\centering
\setlength{\tabcolsep}{4pt}
\caption{\textbf{Long-horizon dexterous tasks (in-distribution).}
Mean progress score (0--100), including the MolmoAct2 baseline (fully-finetuned per task).}
\label{tab:dexterous}
\begin{tabular}{lcccc}
\toprule
 & BIND & ACT & M.\,Tracks & MolmoAct2 \\
\midrule
Teapot (40 demos)   & \textbf{93}  & 2  & 0  & 0 \\
Cup stacking (40)   & \textbf{81}  & 8   & 6   & 0 \\
Spill cleanup (30)  & \textbf{100} & 0  & 3  & 0 \\
Fold towel (35)     & \textbf{97}  & 63 & 12 & 47 \\
\bottomrule
\end{tabular}
\end{table}

\begin{table}[t!]
\centering
\setlength{\tabcolsep}{4pt}
\caption{\textbf{Backbone ablation (in-distribution pick-and-place).}
Same BIND head and training data with the pretrained vision encoder swapped. Progress is the mean 0--100 score; success is the binary pick-and-place rate.}
\label{tab:backbones}
\begin{tabular}{lcc}
\toprule
Backbone   & Progress & Succ.~\% \\
\midrule
\textbf{DINOv3}     & \textbf{98} & \textbf{82} \\
DAv3       & 90 & 88 \\
DynaFlip   & 84 & 80 \\
PaliGemma  & 88 & 78 \\
$\pi_0$    & 78 & 60 \\
\bottomrule
\end{tabular}
\end{table}

\subsection{Scoring Action--Feature Pairs}
\label{sec:scoring}
To provide a score for each candidate action, each candidate action's spatial position is positionally encoded and concatenated with its corresponding image features before scoring. 
While one option is to provide the classifying network with the full 3D positionally-encoded coordinate, in practice, we use just the height dimension of the coordinate, since the projection already determines which pixel ray the candidate lies on. 
Concretely, the sampled features from all views are concatenated with a positional encoding of height and passed through an MLP,
\begin{equation}
\phi(\mathbf{x}) \;=\; \mathrm{MLP}\!\left(\big[\,F_1(\mathbf{u}^{(1)}(\mathbf{x})) \,\big|\, F_2(\mathbf{u}^{(2)}(\mathbf{x})) \,\big|\, \mathrm{sincos}(z)\,\big]\right) \;\in\; \mathbb{R}^{C},
\end{equation}
where the height embedding is a 16-dimensional sin/cos positional encoding~\cite{vaswani2017attention} of the world-height and the MLP is a small few-layer network.

Note that the image encoder is trained end-to-end with the head: we do not freeze the pretrained features. 
We find finetuning to be critical as it enables the features to better implicitly encode robot trajectories.

\subsection{Rotation and Gripper Prediction}
\label{sec:rotgrip}

Rotation and gripper state are similarly predicted as categorical distributions over a number of discretized rotation bins and gripper bins.
To predict the rotation and gripper states for each extracted 3D position keypoint, we index the cross-view fused feature $\phi(\mathbf{x}_v)$ at the ground-truth candidates during training (teacher forcing) or the argmax candidate at inference, flatten these features across the chunk's $T$ timesteps, and pass them through an MLP that outputs the rotation and gripper trajectories for the whole chunk.

\section{Experiments}
\label{sec:experiments}
We evaluate BIND across three axes---data efficiency, robustness to OOD object positions, and robustness to OOD camera viewpoints---and additionally on long-horizon dexterous tasks.
We perform these studies in the real-world as well as simulation for reproducibility and precise definitions of camera and object distributions.
Since BIND is a general action head, we also demonstrate performance with a number of different backbones, including pure-vision backbones as well as vision-language models.

\paragraph{Baselines}
\textit{ACT}~\cite{zhao2023aloha} (CLS$\to$XYZ): the CLS token of the shared backbone is fed to an MLP that regresses end effector coordinates and gripper state directly---the standard global-regression head.
\textit{Motion Tracks} (CLS$\to$UVZ)~\cite{ren2025motiontracks}: a modified version of Motion Tracks that predicts 2D image-space tracks plus ray-height from the same global token and recovers end effector coordinates from them (instead of using two views for triangulation from the global pixel coordinate regression).
\textit{MolmoAct2}~\cite{molmoact2}: a 7B-parameter vision-language-action model, fully-finetuned per task on the same demonstrations.
The controlled simulation study additionally compares against Diffusion Policy~\cite{chi2023diffusionpolicy} and the point-cloud-based DP3~\cite{ze2024dp3}, and the RoboTwin benchmark against their reported ManiFlow~\cite{maniflow2025}.

\paragraph{Setup}
We use YAM arms for experiments with a scene camera as well as a wrist camera. We obtain the camera pose via a Fiducial Exoskeletons~\cite{smith2026fidex} style calibration board.
All methods use a shared DINOv3~\cite{simeoni2025dinov3} ViT-S/16+ backbone, are trained from a single viewpoint with no data augmentation, and are scored by mean task progress (0--100, weighted by outcome category, rather than binary success).

\paragraph{Tasks}
For the core experiments tasks on data-efficiency and robustness, we use a simple pick-and-place task of picking up a cup and placing it on a plate.
For testing more dexterous and long horizon tasks, we use four more tasks:
The teapot task involves taking off the lid of a teapot, placing a teabag in the teapot, and putting the lid back on the teapot; the lid also has a thin grasp point extrusion which requires high precision.
The cup stacking task requires the robot to pick and place a smaller measuring cup into another larger measuring cup in the correct orientation of the larger measuring cup.
The cup-spill task requires the robot to reorient the fallen cup, pick the rag up, and wipe the liquid spill.
The fold-towel is a bi-manual task that requires folding a flat towel cleanly in half.

\begin{figure*}[t!]
\centering
\includegraphics[width=\textwidth]{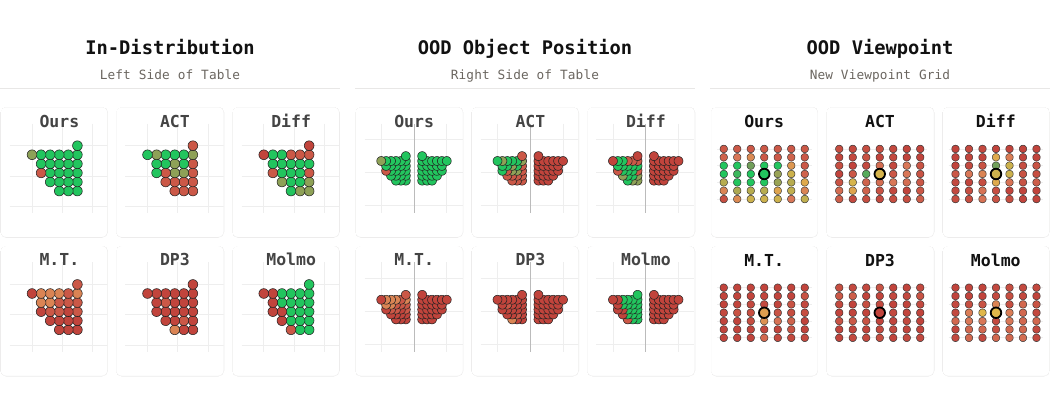}
\vspace{-0.2in}
\caption{\textbf{Controlled simulation experiments (MuJoCo cube pick-and-place).}
We study data efficiency and OOD object position and OOD viewpoints with precise distribution definitions on a controlled MuJoCo cube pick-and-place setup.
We compare BIND against five baselines (Diffusion Policy, MolmoAct2, ACT, Motion Tracks, and DP3).
BIND is the only method that retains meaningful progress under both distribution shifts.}
\label{fig:sim_results}
\end{figure*}

\subsection{Data Efficiency}
\label{sec:data_efficiency}

Our first claim is that BIND yields strong data-efficiency gains. 
We sweep the pick-and-place training set from 3 demonstrations up to 70 (Fig.~\ref{fig:real_results}; Tab.~\ref{tab:data_efficiency}).
BIND sustains a progress score of 86--100 at every training-set size of 5 episodes or more---reaching 100 (10/10 successes) with as few as 5 demonstrations---and degrades to 44 only at 3 episodes.
The ACT baseline never exceeds 36\% at any data size, and Motion Tracks never exceeds 31\%, as we randomize the positions of the cup during training and evaluation.

\subsection{Out-of-Distribution Object Position}
\label{sec:ood_position}
We claim that BIND is less sensitive to the train and test distributions alignment of object positions. 
To demonstrate this, we train the cup place task only on the left half of the table, and then at evaluation move the cup to the right half of the table zero-shot (Fig.~\ref{fig:real_results}; Tab.~\ref{tab:real_ood}).
BIND attains a score of 63 on held-out placements; ACT and Motion Tracks both collapse to 0.
Note that our method is not invariant to the position of the cup---in fact, our method degrades significantly (from 98 in-distribution to 63).
Our claim here is that our method degrades significantly more gracefully and is more \textit{robust} to the out-of-distribution zero-shot setting: while our method does not achieve the same precision as the in-distribution manipulation, when it fails it at least is a `close' failure of reaching near the object, whereas the baselines do not even move towards the correct side of the table. 

\subsection{Out-of-Distribution Camera Viewpoint}
\label{sec:ood_viewpoint}
Our third claim is that BIND is less sensitive to the camera position. We train at just one camera position and evaluate at three different camera positions.
BIND attains a score of 79 aggregated across the three views (Tab.~\ref{tab:real_oodview}); ACT and Motion Tracks collapse to 1 and 4.
Once again, our claim is not of \textit{invariance} to the camera position, but rather that small shifts in camera distribution are of almost no consequence, and extreme camera position changes (e.g. switching to a near birds-eye view) still behave reasonably and even often succeed.

\subsection{Long-Horizon Dexterous Tasks}
\label{sec:dexterous}
We extend beyond simple object pick-and-place to four multi-stage tasks (Fig.~\ref{fig:other_tasks}; Tab.~\ref{tab:dexterous}).
\emph{Teapot} (40 demonstrations): remove the teapot lid, place a teabag inside, and close the lid---BIND 93 vs.\ ACT 2.
\emph{Stack cup} (40 demonstrations): pick the smaller measuring cup and stack it onto the larger measuring cup in the same orientation---BIND 81 vs.\ ACT 8.
\emph{Spill cleanup} (30 demonstrations): reorient a spilled cup upright, pick up a rag, wipe the table---BIND 100 vs.\ ACT 0.
\emph{Fold towel} (35 demonstrations): fold a flat towel cleanly in half with both arms---BIND 97 vs.\ ACT 63.
The fully-finetuned MolmoAct2 scores 0 on the three cup-and-teapot tasks and 47 on fold towel; Motion Tracks never exceeds 12.
Our method succeeds on these more challenging, precise, and long-horizon tasks with a relatively small number of demonstrations (even just 30 examples).

\subsection{Backbone Ablation}
\label{sec:backbones}

Because our action head requires only pixel-aligned feature maps, our method is plug-and-play and the pretrained encoder is swappable (Tab.~\ref{tab:backbones}).
We run the same in-distribution pick-and-place task with five encoders spanning self-supervised, geometric, robotics, VLM, and VLA families~\cite{simeoni2025dinov3,lin2025da3,lee2026dynaflip,beyer2024paligemma,black2024pi0}: DINOv3 82\%, DepthAnything~v3 88\%, DynaFlip 80\%, PaliGemma 78\%, and $\pi_0$ 60\%.
We observe that the spatial features from the general vision encoders (especially DINO) have much qualitatively smoother feature maps than the VLM family of backbones and yield the best results.

\subsection{Controlled Simulation Analysis}
\label{sec:sim}
Measuring the exact object position and camera viewpoint distributions can be challenging in the real-world, so we also offer study in simulation (Tab.~\ref{tab:sim}).
We replicate the two OOD studies in a controlled MuJoCo~\cite{todorov2012mujoco} cube pick-and-place environment where we carefully define the object and camera distributions, comparing against five baselines on identical data: ACT, Diffusion Policy, MolmoAct2, Motion Tracks, and the point-cloud-based DP3 (Fig.~\ref{fig:sim_results}).

\paragraph{Position OOD}
Training positions cover only the left half of the workspace; testing covers held-out mirrored positions on the unseen right half.
In distribution, BIND reaches a score of 94, ahead of the strongest baselines Diffusion Policy (82) and MolmoAct2 (81).
On the held-out half, BIND retains 96 while every baseline drops to 0.

\paragraph{Viewpoint generalization}
We train from a single camera and evaluate zero-shot across a grid of 196 held-out viewpoints.
BIND maintains a score of 49, while all baselines score 12 or lower.
As in the real-world study, BIND is not viewpoint-invariant, but it degrades gracefully where global-regression heads fail outright.

\subsection{LIBERO Benchmark}
\label{sec:libero}

We also evaluate BIND on the LIBERO~\cite{liu2023libero} Spatial and Object suites (Tab.~\ref{tab:libero}).
BIND is trained from scratch on each suite and reaches 92.0\% on Spatial and 94.0\% on Object---above every sub-95 7B baseline on at least one suite and within a few points of the large-scale pretrained frontier.

\subsection{RoboTwin Benchmark}
\label{sec:robotwin}

We also evaluate BIND on five tasks from the RoboTwin bimanual simulation benchmark~\cite{robotwin2025} (Table~\ref{tab:robotwin}).
BIND reaches a mean success rate of 77.8\%, ahead of the strongest baseline, point-cloud-based ManiFlow-3D~\cite{maniflow2025} (61.9), while operating from RGB alone.

\section{Conclusion}
\label{sec:conclusion}

We present BIND, a projection-anchored action formulation that binds each candidate end effector position to the 2D image features at its projection.
In real-world and simulation settings, BIND reaches strong pick-and-place success from as few as 5 demonstrations, is robust to OOD object positions and camera viewpoints where coordinate-regression baselines fail completely, and completes long-horizon dexterous tasks with as few as 30--40 demonstrations.
Controlled simulation experiments replicate both robustness results.

\bibliographystyle{IEEEtran}
\bibliography{refs}

\end{document}